\documentclass[letterpaper]{article}

\makeatletter
\def\input@path{{./}{../AuthorKit27/}}
\makeatother
\usepackage[preprint]{aaai2027}
\usepackage[hyphens]{url}
\usepackage{graphicx}
\usepackage{natbib}
\usepackage{caption}
\usepackage{amsmath}
\usepackage{amssymb}
\usepackage{array}
\usepackage{booktabs}
\usepackage{multirow}
\usepackage{docmute}

\newcommand{\NAV}{\mathrm{NAV}}
\newcommand{\BUILD}{\mathrm{BUILD}}

\title{WikiLoop: Jointly Learning to Build and Navigate Agent-Native Wikis with Downstream Feedback}
\author{
Haoliang Ming, Feifei Li, Wenhui Que
}
\affiliations{
WeChat, Tencent Inc., Beijing, China\\
\texttt{\{hliangming, niyali, victorque\}@tencent.com}
}

\begin{document}


\maketitle

\begin{abstract}
Knowledge-base construction and querying are typically optimized in isolation: retrieval-augmented agents operate over a fixed, externally maintained index, whereas construction receives no signal from downstream use. We present WikiLoop, a feedback-coupled framework that jointly learns to build and navigate an agent-native Wiki, a persistent linked-page knowledge base designed for machine navigation. A role-conditioned shared policy supports two interfaces: a Navigator retrieves evidence from the Wiki to answer queries, and a Builder proposes structured edits evaluated through downstream navigation. The Navigator follows a sufficiency-before-efficiency objective that applies retrieval-cost penalties only after full evidence has been collected. The Builder learns from utility differences: a frozen Navigator scores each candidate edit by its change in downstream performance, while a guard penalty discourages regressions on unrelated queries. Training combines sequential role-specific optimization with a final joint stage over role-homogeneous batches. With Qwen3.5-9B as the common backbone, WikiLoop reaches 62.6 aggregate Answer Correctness on AuthTrace, 6.3 points above LLM-Wiki, base, with the largest gains on multi-document queries. Controlled comparisons support the intended effects of both objectives, and the learned edits remain useful to a held-out Navigator. Paired comparisons indicate that the final shared policy largely retains both role-specific capabilities, improves Navigator and end-to-end Answer Correctness by 0.4 points relative to the corresponding specialist references, and consolidates both interfaces into one model. Without dataset-specific training, WikiLoop also improves over the same-backbone LLM-Wiki, base on HotpotQA and MuSiQue.
\end{abstract}

\section{Introduction}

Retrieval-augmented language models answer knowledge-intensive questions by conditioning generation on external evidence rather than relying only on parametric memory \citep{lewis2020rag,karpukhin2020dpr}. Tool-using agents extend this paradigm by interleaving reasoning with retrieval, allowing later actions to depend on evidence acquired earlier in the trajectory \citep{yao2023react,trivedi2023ircot}. Yet the knowledge artifacts that support these agents are usually produced through a separate construction process. Structured systems may organize corpora into recursive summaries, graphs, or linked Wiki pages, but their construction objectives remain largely separate from feedback received by downstream agents \citep{sarthi2024raptor,edge2024graphrag,ming2026llmwiki}. This separation raises a central question: how should a persistent knowledge base be constructed when its utility is determined by subsequent navigation?

An \emph{agent-native Wiki} is a persistent structured knowledge base designed for machine navigation rather than only for human browsing \citep{ming2026llmwiki}. It organizes corpus information into linked pages that combine synthesized knowledge with provenance to supporting source passages. A Navigator accesses this state through explicit search and read actions and follows inter-page links, whereas a Builder applies schema-constrained patches that create or revise pages, links, and provenance. The Wiki thus serves as both a retrieval substrate and a persistent state whose organization can be optimized through downstream agent feedback.

Jointly learning Wiki construction and navigation raises three optimization challenges. First, adaptive retrieval methods learn whether, when, and how to retrieve \citep{asai2024selfrag,jeong2024adaptiverag,fang2026sparkle}, but their objectives must still balance evidence sufficiency against retrieval efficiency. A uniform cost penalty may favor shorter trajectories that stop before collecting all required evidence. Second, the Builder requires credit for the marginal effect of each edit. Absolute post-edit performance confounds edit quality with the quality of the pre-edit Wiki, whereas unconstrained gains on target queries may degrade unrelated queries. Existing work uses downstream answer utility to train retrieval components \citep{liu2026agenticr} and before/after recoverability to assign credit to memory updates \citep{yan2026himpo}. For a persistent Wiki, however, credit must reflect an isolated edit's marginal downstream utility while accounting for regressions on unrelated queries. Third, the Navigator and Builder use different states, action spaces, and rewards. Although one language model can support multiple retrieval or tool-use roles \citep{zhu2025rolerag,mo2025matpo}, joint optimization may introduce interference, so gains in one role need not preserve the other. A shared policy must coordinate both objectives while retaining each role's specialized capabilities.

We introduce \textbf{WikiLoop}, a feedback-coupled framework that connects Wiki construction to downstream navigation within a role-conditioned shared policy. As shown in Figure~\ref{fig:wikiloop}, joint supervised fine-tuning initializes both roles, followed by Navigator RL, Builder RL with a frozen Navigator evaluator, and Joint RL over interleaved role-homogeneous batches. The Navigator prioritizes evidence sufficiency before retrieval efficiency, while the Builder learns from the downstream utility differences induced by isolated Wiki edits.

Our contributions are threefold:
\begin{enumerate}
    \item \textbf{Feedback-coupled agent-native Wiki learning.} We formulate construction and navigation as a coupled learning problem within one role-conditioned policy and develop a sequential procedure that first specializes and then consolidates the two interfaces. The roles operate over the same structured Wiki and share all trainable parameters while retaining distinct prompts, states, action schemas, and rewards.
    \item \textbf{Role-specific objectives for navigation and construction.} Sufficiency-gated navigation delays cost pressure until full evidence coverage, while guarded utility-difference supervision assigns marginal credit to structured edits. A frozen Navigator scores isolated before/after Wiki states, affected and guard queries separate target benefit from unrelated-query regression, and a held-out Navigator tests transfer beyond the training evaluator.
    \item \textbf{Mechanism-aligned evaluation and empirical evidence.} With Qwen3.5-9B, WikiLoop reaches 62.6 All AC on AuthTrace \citep{wu2026authtrace}, exceeding Joint SFT and LLM-Wiki, base by 4.4 and 6.3 points, respectively. Controlled comparisons support the intended effects of both role-specific objectives. Across the training lineage, Navigator AC rises from 58.2 after Joint SFT to 60.4 after NAV-RL, declines modestly to 59.7 after BUILD-RL, and recovers to 60.8 after Joint RL, which yields a single policy supporting both interfaces. Same-backbone zero-shot gains on HotpotQA and MuSiQue \citep{yang2018hotpotqa,trivedi2022musique}, together with larger-scale results, further assess transfer.
\end{enumerate}

\section{Related Work}

\textbf{Retrieval and agentic knowledge access.} Retrieval-augmented generation and dense passage retrieval established the standard pattern of selecting evidence from an external corpus before generation \citep{lewis2020rag,karpukhin2020dpr}. ReAct and IRCoT extend this pattern to iterative access by interleaving language-model reasoning with retrieval or other actions \citep{yao2023react,trivedi2023ircot}. More recent systems learn retrieval decisions directly: Self-RAG learns retrieval and critique through reflection tokens, Adaptive-RAG routes queries among retrieval strategies based on estimated complexity, and SPARKLE trains a structured policy for adaptive retrieval \citep{asai2024selfrag,jeong2024adaptiverag,fang2026sparkle}. Agentic-R incorporates downstream supervision into retriever learning by combining local relevance with global answer correctness \citep{liu2026agenticr}. These methods optimize evidence access or the retriever itself. WikiLoop instead uses retrieval outcomes to supervise the persistent structured artifact being accessed, while its Navigator objective orders evidence sufficiency before retrieval cost.

\textbf{Memory writing and knowledge editing.} Structured retrieval systems replace flat passage collections with artifacts designed for multi-step access. RAPTOR recursively clusters and summarizes text into a retrieval tree, GraphRAG combines an entity graph with community summaries, and HippoRAG retrieves over graph-structured long-term memory \citep{sarthi2024raptor,edge2024graphrag,gutierrez2024hipporag}. HippoRAG~2 incorporates passages into this memory formulation, while LightRAG supports incremental graph-based indexing \citep{gutierrez2025hipporag2,guo2025lightrag}. LLM-Wiki organizes knowledge into linked, agent-navigable pages and provides the representation adopted by WikiLoop \citep{ming2026llmwiki}. WikiLoop retains this representation but learns structural edits from their marginal downstream utility, connecting artifact construction to subsequent use.

Agent-memory systems treat writing more explicitly as an action. MemGPT manages information across memory tiers, and A-Mem dynamically creates and links structured memories \citep{packer2023memgpt,xu2025amem}. Memory-R1 and AgeMem train policies to manage memory operations through reinforcement learning \citep{yan2025memoryr1,yu2026agemem}. HiMPO estimates the utility of an in-trajectory memory update through before/after recoverability under a shared pre-write state and assigns hindsight-filtered credit to the memory-writing tokens \citep{yan2026himpo}. WikiLoop instead learns schema-constrained edits to a persistent structured Wiki. A frozen downstream Navigator evaluates each candidate patch in isolation, affected and guard queries measure target gains and unrelated-query regressions separately, and a separately trained held-out Navigator tests whether the edits transfer beyond the training evaluator. The two methods therefore share a before/after credit principle but optimize different objects and interfaces: in-trajectory memory-writing tokens in HiMPO versus schema-constrained edits to an external structured Wiki in WikiLoop.

\textbf{Role-conditioned agent learning.} Role conditioning allows one model to support heterogeneous agent behaviors through a shared policy interface. RoleRAG performs role-specific token optimization for multiple RAG functions, while MATPO studies policy optimization across planner and worker roles in tool-integrated models \citep{zhu2025rolerag,mo2025matpo}. WikiLoop conditions a shared policy on navigation or construction over the same persistent Wiki. The roles retain distinct state representations, action spaces, and rewards, while role-homogeneous batches coordinate their optimization.

\section{Method}

\begin{figure*}[t]
\centering
\includegraphics[width=\textwidth]{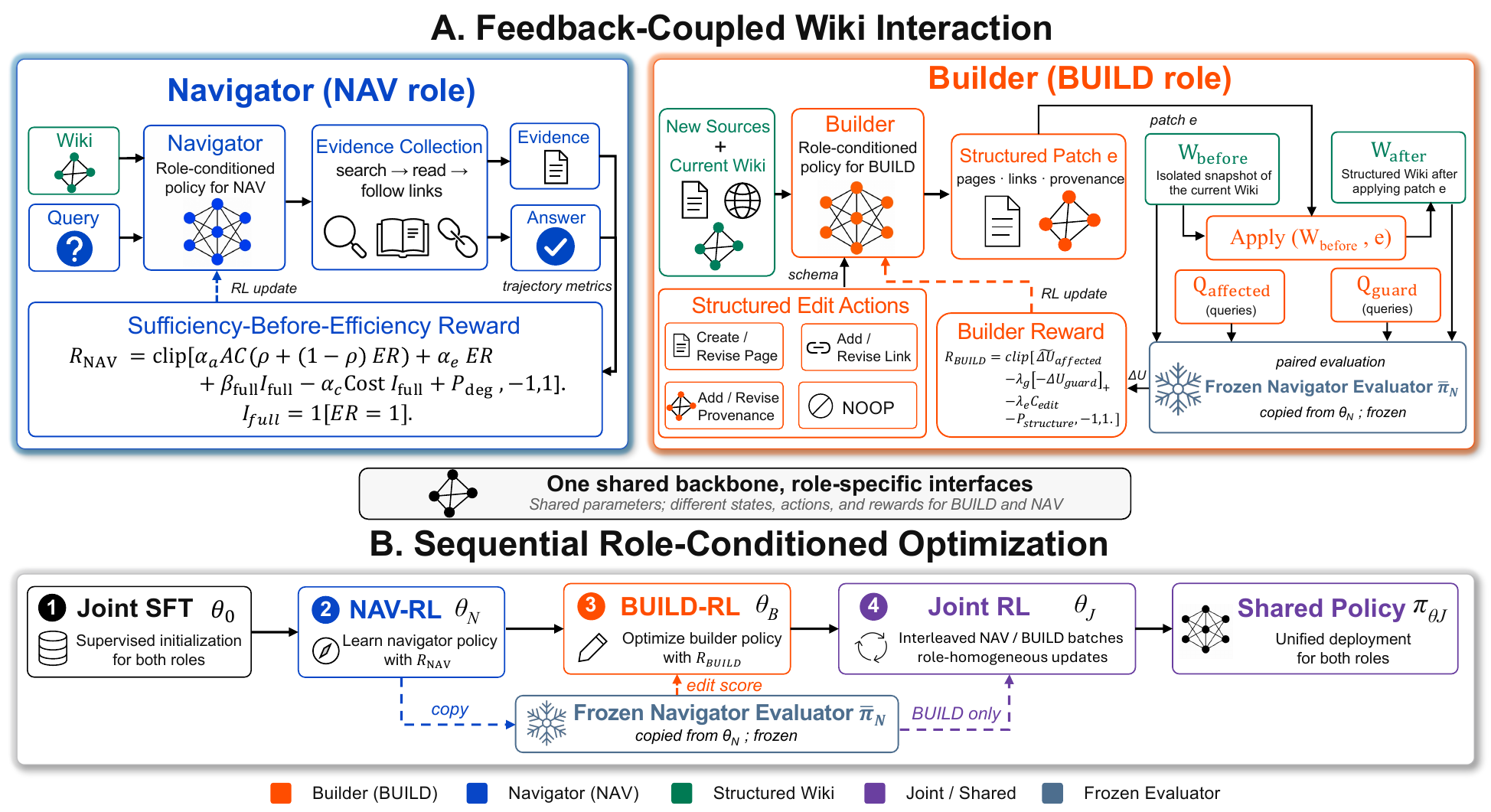}
\caption{Overview of WikiLoop. The Navigator retrieves evidence from a structured Wiki under a sufficiency-before-efficiency objective. The Builder proposes structured patches whose downstream utility is measured by a frozen Navigator on isolated before- and after-edit Wiki states. Sequential NAV-RL and BUILD-RL are followed by Joint RL that interleaves role-homogeneous Navigator and Builder batches, yielding a single role-conditioned shared policy.}
\label{fig:wikiloop}
\end{figure*}

\subsection{Problem Formulation and Framework Overview}

WikiLoop operates on the linked-page Wiki representation introduced by LLM-Wiki \citep{ming2026llmwiki}. A Wiki state $W$ consists of structured pages, inter-page links, and provenance to source documents. Given a query, the \emph{Navigator} issues search and read actions, follows links, accumulates evidence, and returns an answer. Given a new source and the current Wiki state, the \emph{Builder} proposes a schema-constrained structural edit $e$ that creates or revises pages, links, and provenance.

Each candidate edit is paired with an affected set $Q_{\mathrm{affected}}$ containing queries expected to benefit from the edit and a guard set $Q_{\mathrm{guard}}$ used to detect unintended regressions. A frozen Navigator executes both sets on isolated Wiki states before and after applying $e$, then scores the edit by the resulting change in downstream utility. This construction assigns edit-level credit according to the consequences for the downstream agent that consumes the Wiki.

A shared policy represents both roles by conditioning on the active role $r$,
\begin{equation}
\begin{gathered}
\pi_\theta(a_t\mid s_t,r),\\
\pi_\theta^N=\pi_\theta(\cdot\mid r=\NAV),\\
\pi_\theta^B=\pi_\theta(\cdot\mid r=\BUILD).
\end{gathered}
\end{equation}
The Navigator and Builder share model parameters but retain distinct role prompts, state representations, action spaces, and reward functions. Figure~\ref{fig:wikiloop} summarizes the Wiki state, the two role interfaces, the Builder feedback path, and the sequential optimization stages.

Training proceeds sequentially. Joint supervised fine-tuning first produces $\theta_0$, after which the policy evolves as
\begin{equation}
\theta_0
\xrightarrow{\mathrm{NAV\text{-}RL}}\theta_N
\xrightarrow[\bar{\pi}_N\ \mathrm{frozen}]{\mathrm{BUILD\text{-}RL}}\theta_B
\xrightarrow[\bar{\pi}_N\ \mathrm{frozen}]{\mathrm{Joint\ RL}}\theta_J.
\end{equation}
Following NAV-RL, we copy the Navigator interface as $\bar{\pi}_N=\pi_{\theta_N}^{N}$ and keep it fixed throughout BUILD-RL and Joint RL. BUILD-RL produces the Builder-specialized state $\theta_B$, and the subsequent joint stage produces the final role-conditioned shared policy $\pi_{\theta_J}$.

This ordering first provides the Builder with a trained downstream Navigator. Freezing $\bar{\pi}_N$ keeps the utility scale for candidate edits stationary while the Builder changes. The final joint stage then revisits both role objectives to consolidate their interfaces in a single shared policy.

\subsection{Sufficiency-Before-Efficiency Navigation}

The Navigator must balance evidence sufficiency with retrieval efficiency. Penalizing cost before sufficient evidence has been collected can favor prematurely short trajectories, so WikiLoop activates ordinary cost pressure only after full evidence completion. For reward computation, answer correctness $AC$ and evidence recall $ER$ are normalized to $[0,1]$. Let $\widetilde{T}$, $\widetilde{N}_s$, and $\widetilde{N}_c$ denote normalized costs for trajectory tokens, Wiki-search calls, and content-read calls, respectively. We define
\begin{equation}
\mathrm{Cost}=\gamma_T\widetilde{T}
+\gamma_s\widetilde{N}_s
+\gamma_c\widetilde{N}_c,
\end{equation}
where the supplement provides the exact normalizers and weights. Let $I_{\mathrm{full}}=\mathbf{1}[ER=1]$ indicate full evidence completion. The non-positive term $P_{\mathrm{deg}}$ penalizes invalid actions, premature stopping, trajectories with no tool call, and low tool-success rates; its exact conditions and coefficients are specified in the supplement. The clipped trajectory reward is
\begin{equation}
\begin{aligned}
R_{\NAV}=\operatorname{clip}\Big[&
\alpha_a AC\bigl(\rho+(1-\rho)ER\bigr)+\alpha_e ER\\
&+\beta_{\mathrm{full}}I_{\mathrm{full}}\\
&-\alpha_c\mathrm{Cost}\,I_{\mathrm{full}}+P_{\mathrm{deg}},-1,1\Big].
\end{aligned}
\end{equation}
Here $\rho$ sets the minimum fraction of answer credit retained when $ER=0$, and $\beta_{\mathrm{full}}$ weights the full-evidence bonus. The evidence-conditioned factor increases answer credit as more required evidence is recovered, while the separate $ER$ term provides a dense coverage signal. The indicator $I_{\mathrm{full}}$ activates both the completion bonus and the cost penalty.

This gate implements the intended sufficiency-before-efficiency ordering. Before completion, the reward differentiates trajectories through evidence-conditioned answer correctness, evidence recall, and $P_{\mathrm{deg}}$ without applying the ordinary cost penalty. After completion, the bonus rewards full coverage and the cost term favors more economical successful trajectories. The term $P_{\mathrm{deg}}$ continues to discourage degenerate behavior throughout navigation.

\subsection{Utility-Difference Wiki Construction}
\label{sec:builder}

Builder training should attribute credit to the change caused by an individual patch rather than to the absolute quality of the resulting Wiki. Post-edit utility may remain high because $W_{\mathrm{before}}$ was already useful, even when the patch contributes little. We therefore score each candidate patch by its downstream consequences under the frozen evaluator. For query $q$ and Wiki state $W$, we define the navigation utility
\begin{align}
U_q(W;\bar{\pi}_N)={}&w_a AC\bigl(\kappa+(1-\kappa)ER\bigr) \nonumber\\
&+w_e ER-w_c\,\mathrm{Cost},
\end{align}
where $w_a$, $w_e$, and $w_c$ weight answer correctness, evidence recall, and navigation cost, respectively, and $\kappa$ is the minimum fraction of answer utility retained when $ER=0$. We constrain $w_a+w_e+w_c=1$ with $w_a,w_e,w_c\geq0$. The values of $AC$, $ER$, and $\mathrm{Cost}$ are computed from $\bar{\pi}_N$'s trajectory for $q$ on $W$. For a query set $Q$, the mean utility is
\begin{equation}
U(Q,W;\bar{\pi}_N)=\frac{1}{|Q|}\sum_{q\in Q}U_q(W;\bar{\pi}_N).
\end{equation}
Each training instance contains up to three affected queries selected by page/source overlap or semantic relevance and two same-author guard queries without page or source overlap; fallback statistics appear in the supplement.

Each rollout evaluates one patch on an isolated Wiki copy. Starting from $W_{\mathrm{before}}$, we apply a valid patch $e$ to obtain $W_{\mathrm{after}}=\operatorname{Apply}(W_{\mathrm{before}},e)$, then run the frozen Navigator on both states. For either query set $Q\in\{Q_{\mathrm{affected}},Q_{\mathrm{guard}}\}$, the utility difference is
\begin{equation}
\begin{aligned}
\Delta U_Q(e)={}&U(Q,W_{\mathrm{after}};\bar{\pi}_N)\\
&-U(Q,W_{\mathrm{before}};\bar{\pi}_N).
\end{aligned}
\end{equation}
We denote the two quantities by $\Delta U_{\mathrm{affected}}$ and $\Delta U_{\mathrm{guard}}$. Because $U$ averages query-level utility, each difference is the arithmetic mean of $U_q(W_{\mathrm{after}})-U_q(W_{\mathrm{before}})$ over its query set.

The candidate state is discarded after scoring rather than committed to the training Wiki. A patch that cannot be parsed, violates the Wiki schema, invokes an inapplicable operation, or fails during application receives $R_{\BUILD}=-1$. A valid \textsc{noop} that produces no state change receives $R_{\BUILD}=0$ and bypasses edit-cost and structural penalties. Let $[x]_+=\max(0,x)$ and $\widetilde{\Delta U}_{\mathrm{affected}}=\operatorname{clip}(\Delta U_{\mathrm{affected}},-1,1)$. For a valid, non-empty edit, the reward is
\begin{equation}
\begin{aligned}
R_{\BUILD}=\operatorname{clip}\Big[&
\widetilde{\Delta U}_{\mathrm{affected}}
-\lambda_g[-\Delta U_{\mathrm{guard}}]_+\\
&-\lambda_e C_{\mathrm{edit}}
-P_{\mathrm{structure}},-1,1\Big].
\end{aligned}
\end{equation}
The positive-part term makes guard protection one-sided: negative guard utility changes are penalized, whereas positive changes receive no additional reward. Here $C_{\mathrm{edit}}\in[0,1]$ is the normalized edit size, and $P_{\mathrm{structure}}$ aggregates penalties for orphaning, structural complexity from page growth, fragmentation, and overlength. The supplement specifies the normalizer, coefficients, and component definitions.

Paired before/after evaluation isolates the candidate's marginal effect relative to the same starting state. Freezing $\bar{\pi}_N$ keeps the utility scale fixed as the Builder changes, avoiding a moving reward criterion caused by actor--evaluator co-adaptation. Together, these choices assign edit-level credit under a stable downstream consumer.

\subsection{Role-Conditioned Joint Optimization}

Although the Navigator and Builder expose different action interfaces, they rely on substantially overlapping capabilities over the same Wiki. Both roles must identify task-relevant information, understand page and link structure, track provenance, and reason about evidence requirements. The Navigator uses these capabilities to locate and compose evidence, whereas the Builder uses them to predict how structural edits will affect future navigation. They therefore provide complementary read and write views of the same knowledge environment, motivating shared model parameters.

The roles nevertheless require different credit signals. Role-specific stages adapt each interface using its own action space and reward. Because $\theta_B$ is most recently optimized for Wiki construction, sequential training may shift the navigation behavior learned during NAV-RL. We therefore initialize a final joint stage from $\theta_B$ to revisit both objectives and consolidate the interfaces into one shared policy.

Role conditioning preserves distinct interaction protocols within the shared parameterization. The active role determines the prompt, observation serialization, action schema, and reward function, while all trainable parameters are shared. Let $p(r)$ denote the role-sampling distribution. The joint objective is
\begin{equation}
\begin{aligned}
\max_\theta\quad &\mathbb{E}_{r\sim p(r)}
\mathbb{E}_{\tau\sim\pi_\theta(\cdot\mid r)}
\left[R_r(\tau)\right], \\
R_r(\tau) &=\begin{cases}
R_{\NAV}(\tau), & r=\NAV,\\
R_{\BUILD}(\tau;\bar{\pi}_N), & r=\BUILD.
\end{cases}
\end{aligned}
\end{equation}
This objective coordinates two role-specific learning signals within a shared parameter space.

Navigator and Builder trajectories are not directly comparable because they differ in action spaces, trajectory structures, and reward definitions. Group-relative optimization constructs advantages through comparisons within each rollout group; mixing roles would therefore entangle role identity with within-role trajectory quality. WikiLoop instead interleaves role-homogeneous batches. A Navigator batch contains only navigation trajectories scored by $R_{\NAV}$, whereas a Builder batch contains only candidate-patch trajectories scored by $R_{\BUILD}$.

Builder batches continue to use the frozen evaluator $\bar{\pi}_N$. The fixed downstream consumer keeps marginal edit utility comparable across BUILD-RL and Joint RL while the shared actor changes.

For each role-homogeneous batch, WikiLoop computes group-relative advantages following GRPO \citep{shao2024deepseekmath} and applies the GSPO sequence-level policy objective \citep{zheng2025gspo}. Joint RL reuses the role-specific rewards and coefficients without modification. Updates from both roles modify the same parameter vector, producing the final policy $\pi_{\theta_J}$. At inference time, the role prompt selects the Navigator or Builder interface without changing model parameters.

\section{Experiments}

We organize the evaluation around four questions: (1) whether WikiLoop improves end-to-end performance under a shared backbone; (2) whether the Navigator and Builder objectives produce their intended behaviors; (3) whether joint optimization retains both role-specific capabilities; and (4) whether the observed gains transfer across model scales and datasets.

\subsection{Experimental Setup}
\label{sec:setup}

\textbf{Datasets and metrics.} AuthTrace contains 2,099 questions from 860 public-domain works by five Chinese prose writers \citep{wu2026authtrace}: 1,285 Single ($f=1$), 552 Low ($f=2$), and 262 High ($f\geq3$), where $f$ is the number of required documents. Aggregate AC uses the official stratum counts. Tables report AC on $[0,100]$; rewards use AC and Evidence Recall (ER) on $[0,1]$. Role-specific metrics are defined with their experiments below.

\textbf{Models and baselines.} Primary experiments use Qwen3.5-9B as the common answer backbone and share the AuthTrace corpus, queries, and rubric. Baselines are closed-book inference, Vanilla RAG (BM25 and Dense), RAPTOR \citep{sarthi2024raptor}, LightRAG \citep{guo2025lightrag}, HippoRAG~2 \citep{gutierrez2025hipporag2}, and LLM-Wiki, base \citep{ming2026llmwiki}. Public systems retain their recommended indexing and retrieval procedures, making these system-level comparisons. We also evaluate Qwen3.5-35B-A3B and transfer to HotpotQA and MuSiQue \citep{yang2018hotpotqa,trivedi2022musique}.

\textbf{Training and calibration.} We construct synthetic NAV and BUILD data from fixed document-level partitions, using DeepSeek-V4-Flash as the teacher; all AuthTrace questions are reserved for final evaluation. We select Navigator coefficients on NAV validation, freeze the resulting evaluator, and select $(w_a,w_e,w_c,\kappa)$ on Builder validation. Evaluation data are excluded from supervision, rewards, filtering, calibration, and model selection. Supplementary Tables~S1--S2 detail optimization, data construction, and split integrity.

\textbf{Evaluation protocol.} Builder comparisons fix initialization, data, evaluator, rollout budget, optimizer, and model selection. A different-seed NAV-RL run supplies the held-out Navigator for final evaluation. Following AuthTrace, DeepSeek-V4-Flash scores anonymized outputs at temperature zero with a fixed 0--3 rubric. GLM-5.1 and blinded human evaluation preserve the ranking, yielding WikiLoop 62.5/62.8 All AC (Supplementary Tables~S4/S7).

\subsection{Same-Backbone End-to-End Performance}

We compare WikiLoop with closed-book inference, sparse and dense RAG, structured-retrieval systems, and LLM-Wiki, base under a shared answer backbone. Single, Low, and High stratify queries by required-document count; LLM-Wiki, base provides the most direct comparison within the same Wiki-building framework.

\begin{table}[t]
\centering
\small
\setlength{\tabcolsep}{3.6pt}
\begin{tabular*}{\columnwidth}{@{\extracolsep{\fill}}lcccc@{}}
\toprule
Method & Single & Low & High & All \\
\midrule
None (Closed-book) & 2.6 & 4.4 & 3.4 & 3.2 \\
Vanilla RAG (BM25) & 43.8 & 12.0 & 8.0 & 31.0 \\
Vanilla RAG (Dense) & 41.5 & 12.9 & 11.6 & 30.2 \\
RAPTOR & 39.2 & 10.6 & 9.8 & 28.0 \\
LightRAG & 42.6 & 8.1 & 4.9 & 28.8 \\
HippoRAG~2 & \textbf{69.8} & 36.5 & 28.0 & 55.8 \\
LLM-Wiki, base & 66.7 & 42.0 & 35.2 & 56.3 \\
\textbf{WikiLoop} & 69.1 & \textbf{54.8} & \textbf{47.5} & \textbf{62.6} \\
\bottomrule
\end{tabular*}
\caption{System-level AuthTrace AC ($\uparrow$) under a shared Qwen3.5-9B answer backbone; methods retain their native indexing and retrieval procedures.}
\label{tab:main}
\end{table}

WikiLoop reaches 62.6 All AC (Table~\ref{tab:main}), exceeding LLM-Wiki, base and HippoRAG~2 by 6.3 and 6.8 points. Relative to LLM-Wiki, base, its gains are 2.4/12.8/12.3 on Single/Low/High. WikiLoop remains competitive with HippoRAG~2 on Single (69.1 versus 69.8) and performs best on Low and High, concentrating its advantage on multi-document queries.

Paired tests find reliable gains over LLM-Wiki, base in every stratum and over HippoRAG~2 on All, Low, and High. HippoRAG~2's $0.7$-point advantage on Single is not statistically reliable (Supplementary Table~S8).

\subsection{Analysis of Role-Specific Objectives}
\label{sec:role-objectives}

\textbf{Navigator objective.} We compare Joint SFT, ungated NAV-RL, and full NAV-RL. The matched ablation changes only the cost term, replacing $-\alpha_c\mathrm{Cost}\,I_{\mathrm{full}}$ with $-\alpha_c\mathrm{Cost}$. Premature Stop is active termination with $I_{\mathrm{full}}=0$; Cost@Full averages cost only over trajectories with $I_{\mathrm{full}}=1$.

\textbf{Builder objective.} We compare Joint SFT, absolute post-edit utility, unguarded utility difference, and full BUILD-RL. This sequence separates the effects of marginal credit assignment and guard protection. Affected $\Delta AC$ measures correctness changes on affected queries; In-Eval. and Held-out Navigator $\Delta U$ use the frozen training evaluator and the separately trained, different-seed Navigator, respectively.

For evaluation, let $\Delta U_q(e)=U_q(W_{\mathrm{after}})-U_q(W_{\mathrm{before}})$. We report query-level guard regression,
\begin{equation}
\mathrm{GuardReg}(e)=
\frac{1}{|Q_{\mathrm{guard}}|}
\sum_{q\in Q_{\mathrm{guard}}}\max(0,-\Delta U_q(e)).
\end{equation}
Unlike the set-level training penalty, this query-level metric does not allow an improvement on one guard query to offset a regression on another.

\begin{table}[t]
\centering
\small
\renewcommand{\arraystretch}{1.04}
\setlength{\tabcolsep}{2.4pt}
\begin{tabular*}{\columnwidth}{@{\extracolsep{\fill}}lrrrr@{}}
\multicolumn{5}{c}{\textbf{A. Navigator Objective}}\\[0.25ex]
\toprule
Objective & AC $\uparrow$ & ER Full $\uparrow$ & Prem. Stop $\downarrow$ & Cost@Full $\downarrow$ \\
\midrule
Joint SFT & 58.2 & 66.8 & 18.9 & 0.411 \\
NAV-RL (Ungated) & 58.1 & 64.0 & 24.2 & \textbf{0.364} \\
\textbf{NAV-RL (Full)} & \textbf{60.4} & \textbf{76.1} & \textbf{11.1} & 0.379 \\
\bottomrule
\end{tabular*}

\par\vspace{0.8ex}

\begin{tabular*}{\columnwidth}{@{\extracolsep{\fill}}lcccc@{}}
\multicolumn{5}{c}{\textbf{B. Builder Objective}}\\[0.25ex]
\toprule
Objective & Aff. $\uparrow$ & In-Eval. $\uparrow$ & Held-out $\uparrow$ & Guard $\downarrow$ \\
\midrule
Joint SFT & 4.9 & 0.040 & 0.026 & 0.021 \\
Absolute $U_{\mathrm{after}}$ & 6.2 & 0.054 & 0.034 & 0.025 \\
$\Delta U$, No Guard & \textbf{12.8} & \textbf{0.118} & 0.077 & 0.046 \\
\textbf{BUILD-RL (Full)} & 12.1 & 0.111 & \textbf{0.084} & \textbf{0.016} \\
\bottomrule
\end{tabular*}
\caption{Analysis of the role-specific objectives. Panel A compares Joint SFT, ungated NAV-RL, and full NAV-RL; ER Full and Premature Stop are percentages. In Panel B, Aff. is affected-query $\Delta AC$, In-Eval. and Held-out are $\Delta U$, and Guard is Guard Regression.}
\label{tab:objectives}
\end{table}

\textbf{Navigator results.} Relative to Joint SFT, full NAV-RL raises AC from 58.2 to 60.4 and full-evidence completion from 66.8\% to 76.1\%, while reducing premature stopping from 18.9\% to 11.1\% and Cost@Full from 0.411 to 0.379. The gain therefore does not arise solely from longer retrieval: among successful full-evidence trajectories, the learned policy is also less costly than its supervised initialization.

Ungated NAV-RL attains the lowest Cost@Full (0.364), but AC falls to 58.1, full-evidence completion to 64.0\%, and premature stopping rises to 24.2\%. Relative to this ablation, full NAV-RL gains 2.3 AC and 12.1 points in full-evidence completion while reducing premature stopping by 13.1 points. Its 0.015 higher cost among full-evidence trajectories is thus accompanied by substantially better evidence completion, supporting the sufficiency-before-efficiency design.

\textbf{Builder results.} Absolute post-edit utility yields modest gains over Joint SFT: affected $\Delta AC$ rises from 4.9 to 6.2, In-Eval. $\Delta U$ from 0.040 to 0.054, and held-out Navigator $\Delta U$ from 0.026 to 0.034. Unguarded $\Delta U$ raises the same metrics to 12.8, 0.118, and 0.077. This contrast supports paired before/after utility as a more informative credit signal for an individual patch than its absolute post-edit state.

The stronger affected-query gain without guard protection comes with higher Guard Regression (0.046). Adding the guard term reduces this value to 0.016, a 65.2\% relative reduction, while affected $\Delta AC$ changes only from 12.8 to 12.1. In-Eval. $\Delta U$ similarly changes from 0.118 to 0.111, whereas held-out Navigator $\Delta U$ increases from 0.077 to 0.084. The guard term therefore limits regressions on unrelated queries while retaining most target-query benefit; the positive held-out result further indicates that the edits remain useful to a separately trained Navigator excluded from Builder training and model selection.

Paired tests further support both mechanisms. Sufficiency gating improves AC, full-evidence completion, and premature stopping; marginal utility improves held-out edit utility, and guard protection reduces Guard Regression. By contrast, the increase in Guard Regression from absolute to unguarded marginal utility and the small held-out utility change from the unguarded to the full objective are not statistically reliable (Supplementary Table~S8).

Supplementary Figure~S1 shows that role-specific RL primarily reduces premature stopping (18.9\% to 11.1\%) and evidence coverage gaps (28.9\% to 17.6\%); other error categories improve more modestly.

\subsection{Capability Retention in the Shared Policy}
\label{sec:shared-policy}

We trace the NAV interface across sequential training under a fixed Wiki and evaluation protocol, then compare the final shared policy with the retained role specialists and their end-to-end composition.

\begin{table*}[t]
\centering
\small
\setlength{\tabcolsep}{3.6pt}
\begin{tabular*}{\textwidth}{@{\extracolsep{\fill}}lcccccc@{}}
\toprule
Evaluation & Joint SFT & Specialist Ref. & Post-BUILD NAV & Joint
& $\Delta$ (95\% CI) & Holm $p$ \\
\midrule
Navigator AC $\uparrow$
& 58.2 & 60.4 & 59.7 & \textbf{60.8}
& $\boldsymbol{+0.4\;[+0.08,\,+0.73]}$ & \textbf{0.034} \\
Navigator ER Full $\uparrow$
& 66.8 & \textbf{76.1} & 74.5 & 75.9
& N/A & N/A \\
Navigator Prem. Stop $\downarrow$
& 18.9 & 11.1 & 12.5 & \textbf{10.9}
& N/A & N/A \\
Builder held-out Nav. $\Delta U$ $\uparrow$
& 0.026 & \textbf{0.084} & N/A & 0.081
& $-0.003\;[-0.008,\,+0.002]$ & 0.231 \\
End-to-end All AC $\uparrow$
& 58.2 & 62.2 & N/A & \textbf{62.6}
& $\boldsymbol{+0.4\;[+0.12,\,+0.69]}$ & \textbf{0.018} \\
\bottomrule
\end{tabular*}
\caption{Capability evolution and retention. Navigator rows trace Joint SFT, NAV-RL (Specialist Ref.), post-BUILD-RL NAV, and Joint RL. For Builder/end-to-end evaluation, Specialist Ref. denotes BUILD-RL/the specialist pair. $\Delta$ is Joint minus Specialist Ref.; CIs and Holm-adjusted $p$-values are reported for the three primary comparisons only. N/A marks inapplicable cells.}
\label{tab:joint}
\end{table*}

Table~\ref{tab:joint} shows specialization, temporary drift, and recovery. NAV-RL raises AC from 58.2 to 60.4 and ER Full from 66.8 to 76.1 while reducing premature stopping from 18.9 to 11.1. After BUILD-RL, these metrics shift modestly to 59.7, 74.5, and 12.5; Joint RL then reaches 60.8 AC, 75.9 ER Full, and 10.9 premature stopping.

Relative to the specialists, the shared policy improves Navigator and end-to-end AC by 0.4 points each, with paired-bootstrap intervals excluding zero and Holm-adjusted $p$-values below 0.05. Builder utility changes from 0.084 to 0.081 ($\Delta=-0.003$), with a CI spanning zero ($p=0.231$). Joint RL therefore recovers the temporary Navigator drift while retaining Builder utility within the uncertainty of the specialist comparison.

Supplementary Tables~S5--S6 report training allocation and system cost. Relative to LLM-Wiki, base, WikiLoop combines a 6.3-point All AC gain with lower online latency (16.90 to 15.76 seconds per query) and lower full-Wiki construction cost (194.00M to 185.15M tokens). Within the same Wiki-building framework, these results indicate a better quality--cost trade-off.

\subsection{Cross-Dataset Transfer}

We evaluate zero-shot transfer to HotpotQA and MuSiQue without target-dataset training or calibration. Table~\ref{tab:cross-dataset} separates contextual GLM-5.1 references from the controlled Qwen3.5-9B comparison between WikiLoop and LLM-Wiki, base.

\begin{table}[t]
\centering
\small
\setlength{\tabcolsep}{2.4pt}
\begin{tabular*}{\columnwidth}{@{\extracolsep{\fill}}lrrrr@{}}
\multicolumn{5}{c}{\textbf{A. GLM-5.1 Contextual References}}\\[0.2ex]
\toprule
\multirow{2}{*}{Method} & \multicolumn{2}{c}{HotpotQA} & \multicolumn{2}{c}{MuSiQue} \\
\cmidrule(lr){2-3}\cmidrule(lr){4-5}
& F1 $\uparrow$ & EM $\uparrow$ & F1 $\uparrow$ & EM $\uparrow$ \\
\midrule
None (Closed-book) & 0.551 & 0.442 & 0.456 & 0.372 \\
Vanilla RAG (BM25) & 0.717 & 0.590 & 0.545 & 0.442 \\
Vanilla RAG (Dense) & 0.764 & 0.642 & 0.611 & 0.500 \\
RAPTOR & 0.801 & 0.674 & 0.522 & 0.442 \\
LightRAG & 0.819 & 0.682 & 0.659 & 0.550 \\
HippoRAG~2 & 0.805 & 0.668 & 0.624 & 0.514 \\
LLM-Wiki & 0.839 & 0.710 & 0.739 & 0.634 \\
\bottomrule
\end{tabular*}

\par\vspace{0.55ex}

\begin{tabular*}{\columnwidth}{@{\extracolsep{\fill}}lrrrr@{}}
\multicolumn{5}{c}{\textbf{B. Qwen3.5-9B Controlled Comparison}}\\[0.2ex]
\toprule
\multirow{2}{*}{Method} & \multicolumn{2}{c}{HotpotQA} & \multicolumn{2}{c}{MuSiQue} \\
\cmidrule(lr){2-3}\cmidrule(lr){4-5}
& F1 $\uparrow$ & EM $\uparrow$ & F1 $\uparrow$ & EM $\uparrow$ \\
\midrule
LLM-Wiki, base & 0.752 & 0.584 & 0.579 & 0.438 \\
\textbf{WikiLoop} & \textbf{0.777} & \textbf{0.618} & \textbf{0.614} & \textbf{0.482} \\
\bottomrule
\end{tabular*}
\caption{Zero-shot transfer to HotpotQA and MuSiQue. Panel A reports contextual GLM-5.1 results, whereas Panel B provides the controlled Qwen3.5-9B comparison.}
\label{tab:cross-dataset}
\end{table}

Under Qwen3.5-9B, WikiLoop improves over LLM-Wiki, base on all four controlled metrics. F1/EM increase by 0.025/0.034 on HotpotQA \citep{yang2018hotpotqa} and by 0.035/0.044 on MuSiQue \citep{trivedi2022musique}, corresponding to relative gains of 3.3/5.8\% and 6.0/10.0\%, respectively. The improvements are larger on MuSiQue for both metrics, with the largest absolute gain on MuSiQue EM. Because neither target dataset participates in training or calibration, these results support zero-shot transfer beyond AuthTrace to new corpora and query distributions. The GLM-5.1 block provides protocol context and is not used for same-backbone attribution.

\subsection{Scale Transfer}

We next test whether the benefit of feedback-coupled training transfers from Qwen3.5-9B to Qwen3.5-35B-A3B; Supplementary Table~S3 reports the complete results. The larger backbone raises All AC by 2.5 points for Joint SFT (58.2 to 60.7) and by 2.2 points for WikiLoop (62.6 to 64.8). Within WikiLoop, Single, Low, and High AC increase by 1.7, 3.3, and 1.9 points, respectively.

More importantly, WikiLoop remains effective relative to the supervised initialization at both scales: All AC improves over Joint SFT by 4.4 points at 9B and 4.1 points at 35B-A3B. At both scales, the advantage is larger on multi-document queries, reaching 5.5/7.7 points on Low/High at 9B and 6.1/7.2 points at 35B-A3B, compared with 3.3 and 2.6 points on Single. Increasing backbone capacity benefits Low most within WikiLoop, whereas feedback-coupled training produces its largest advantage over Joint SFT on High. The result establishes transfer to one larger sparse backbone; broader model families and scales remain to be studied.

\section{Limitations}

Candidate patches are evaluated on isolated Wiki copies and discarded after scoring. Fixing the pre-edit state and evaluator isolates the marginal utility of each patch, but it also means that Joint RL does not optimize persistent construction and subsequent navigation within a single closed-loop trajectory. Such a setting would introduce history-dependent credit, a moving evaluation target, and editing errors that propagate into later navigation. Persistent write-back and long-horizon Builder--Navigator co-adaptation therefore remain open directions.

AuthTrace training and evaluation share the underlying corpus and initial Wiki, making the setting transductive at the knowledge-base level despite strict query and annotation isolation. Cross-dataset evaluation covers two benchmarks, and scale transfer covers one larger sparse backbone. Evaluating independently initialized Wikis across broader corpora, domains, languages, and model families would further test the framework's generality.

\section{Conclusion}

WikiLoop connects knowledge construction to downstream navigation by learning Navigator and Builder interfaces within one role-conditioned policy. Its Navigator prioritizes evidence sufficiency before retrieval efficiency, while its Builder learns from guarded before/after utility measured by a frozen Navigator on isolated Wiki states. Experiments on AuthTrace show the largest gains on multi-document queries, and controlled analyses connect these gains to improved evidence completion, marginal edit credit, and protection against regressions on unrelated queries. The learned edits remain useful to a held-out Navigator, while the final shared policy largely retains both specialist capabilities. Improvements also transfer zero-shot to HotpotQA and MuSiQue and persist with a larger sparse backbone. These results establish downstream navigation utility as an effective training signal for structured Wiki construction while supporting navigation and construction within one role-conditioned shared policy.

\bibliography{wikiloop}

\appendix
\setcounter{section}{0}
\setcounter{table}{0}
\setcounter{figure}{0}
\renewcommand{\thetable}{S\arabic{table}}
\renewcommand{\thefigure}{S\arabic{figure}}
\providecommand{\maketitle}{}
\renewcommand{\maketitle}{}

\maketitle

\section{Experimental Details}

\subsection{Implementation and Optimization}

Unless stated otherwise, all primary experiments use Qwen3.5-9B. Table~\ref{tab:supp-training-config} summarizes stage-specific optimization, and Table~\ref{tab:supp-training-cost} reports GPU allocation and wall-clock time.

\paragraph{Randomness and runs.}
Unless stated otherwise, each reported configuration is trained and evaluated once. Each run uses the fixed random seed recorded in its submitted configuration for model initialization, data shuffling, and rollout sampling. The held-out Navigator is trained with a distinct fixed NAV-RL seed from the frozen Navigator evaluator used for Builder rewards.

\begin{table}[!ht]
\centering
\small
\setlength{\tabcolsep}{4pt}
\begin{tabular}{@{}l>{\raggedright\arraybackslash}p{0.69\columnwidth}@{}}
\toprule
Stage & Configuration \\
\midrule
Joint SFT & Init.: Qwen3.5-9B; LR: $1\!\times\!10^{-6}$; updates: 839; batch: 16. \\
NAV-RL (Full) & Init.: Joint SFT; LR: $5\!\times\!10^{-7}$; updates: 120; batch/group: 128/4. \\
BUILD-RL (Full) & Init.: NAV-RL (Full); LR: $5\!\times\!10^{-8}$; updates: 90; batch/group: 32/4. \\
Joint RL & Init.: BUILD-RL (Full); LR: $5\!\times\!10^{-8}$; updates: 90; batch/group: 32/4. \\
\bottomrule
\end{tabular}
\caption{Training configurations. Batch/group denotes batch size and rollouts per prompt.}
\label{tab:supp-training-config}
\end{table}

Validation-based calibration selects $\alpha_a=0.45$, $\alpha_e=0.40$, and $\alpha_c=0.10$ for the Navigator. For the Builder, we compare 15 candidate settings using affected-query improvement, guard regression, and navigation cost, and select $w_a=0.60$, $w_e=0.35$, $w_c=0.05$, and $\kappa=0.2$. The selected coefficients remain fixed throughout BUILD-RL (Full) and Joint RL.

Calibration follows the sequential training pipeline. We first select the Navigator coefficients and freeze the resulting evaluator. We then remove Pareto-dominated Builder settings based on short-horizon pilots and select the final compromise from the remaining candidates. The held-out Navigator is obtained from a separate NAV-RL run under the same setup but with a different random seed. It is reserved for final evaluation and excluded from Builder rewards, calibration, and model selection. All calibration and model selection use validation data exclusively; the AuthTrace evaluation set is used only for final evaluation. Policy optimization uses \texttt{seq-mean-token-mean} loss aggregation, a KL-loss coefficient of $0.003$, and asymmetric clipping widths of $0.20$ and $0.28$.

Joint RL repeats two Navigator batches followed by one Builder batch, with every batch remaining role homogeneous. The resulting $2{:}1$ ratio describes optimizer-update frequency only; because sequence lengths and gradient magnitudes differ across roles, it does not specify their relative token volume or gradient contribution. This Navigator-weighted schedule revisits navigation after initialization from the Builder-specialized state $\theta_B$ while maintaining periodic Builder updates. The schedule is fixed across all Joint RL experiments.

\textbf{Reward implementation.} The Navigator uses $\rho=0.1$ and $\beta_{\mathrm{full}}=0.25$. The main text defines full evidence as $ER=1$; implementation uses the numerically equivalent tolerance $ER\geq1-10^{-9}$. The normalized cost components are
\begin{align}
\widetilde{T} &= \min(T/8000,1), \nonumber\\
\widetilde{N}_s &= \min(N_s/12,1), \nonumber\\
\widetilde{N}_c &= \min(N_c/12,1), \nonumber\\
\mathrm{Cost} &= 0.6\widetilde{T}+0.2\widetilde{N}_s
+0.2\widetilde{N}_c.
\end{align}
where $T$ is the trajectory token count and $N_s$ and $N_c$ count Wiki-search and content-read calls.

\paragraph{Anti-degeneracy shaping.}
The Navigator additionally receives
\begin{equation}
\begin{aligned}
P_{\mathrm{deg}}
={}&-0.6\,\mathbb{I}[N_{\mathrm{tool}}=0]
-0.6\,\mathbb{I}[N_{\mathrm{read}}=0]\\
&-0.3\,\mathbb{I}[N_{\mathrm{search}}=0]
-0.3\,\mathbb{I}[ER=0]\\
&-0.5(1-ER)\,
\mathbb{I}[\mathrm{STOP}\land ER<1]\\
&-0.5\,\mathbb{I}[\mathrm{invalid}]
-0.3\,\mathbb{I}[s_{\mathrm{tool}}<0.5]\\
&-0.5\,\mathbb{I}[N_{\mathrm{resp}}<32
\land N_{\mathrm{read}}=0].
\end{aligned}
\end{equation}
$N_{\mathrm{tool}}$, $N_{\mathrm{read}}$, and
$N_{\mathrm{search}}$ denote tool, read, and search calls;
$s_{\mathrm{tool}}$ is the tool success rate and
$N_{\mathrm{resp}}$ is the response length in tokens.
Multiple conditions may apply simultaneously. We add the resulting penalty before clipping the final reward to $[-1,1]$.

For the Builder, $\lambda_g=0.25$ and $\lambda_e=0.03$. Edit cost is implemented as
\begin{align}
C_{\mathrm{edit}}
&=\min\!\left(\frac{N_{\mathrm{added\ chars}}}{3000},1\right).
\end{align}

\paragraph{Structural shaping.}
For a valid Builder patch, we use
\begin{equation}
\begin{aligned}
P_{\mathrm{structure}}
={}&0.05\,r_{\mathrm{orphan}}
+0.01\,s_{\mathrm{growth}}\\
&+0.03\,r_{\mathrm{frag}}
+0.02\,r_{\mathrm{overlength}},
\end{aligned}
\end{equation}
where
\begin{equation}
s_{\mathrm{growth}}
=
\min\left(
\frac{\max(\Delta_{\mathrm{pages}},0)}{3},1
\right).
\end{equation}
Here, $r_{\mathrm{orphan}}$ and $r_{\mathrm{frag}}$ are the fractions of newly created pages with no incoming Wiki link and fewer than 200 body characters, respectively. $r_{\mathrm{overlength}}$ is the fraction of patch contents longer than 8,000 characters, and $\Delta_{\mathrm{pages}}$ is the increase in page count.

\subsection{Synthetic Training Data and Split Integrity}

We construct all training data from fixed document-level partitions before model training. DeepSeek-V4-Flash serves as the teacher for query generation, navigation demonstrations, and Wiki-construction trajectories. The original 2,099 AuthTrace questions are reserved for final evaluation and are not used during synthetic-data generation. Table~\ref{tab:supp-data} summarizes the resulting training collections.

\begin{table}[!htbp]
\centering
\small
\setlength{\tabcolsep}{2.2pt}
\begin{tabular}{@{}lccc@{}}
\toprule
Collection & Train & Val. & Total / Composition \\
\midrule
NAV-RL queries & 1,976 & 197 & 2,173 \\
BUILD-SFT instances & 2,597 & 374 & 2,971 \\
Joint SFT sequences & N/A & N/A & 11,693 \\
Builder candidates & N/A & N/A & 628 / 2,385 / 335 \\
AuthTrace benchmark & N/A & N/A & 2,099 eval. \\
\bottomrule
\end{tabular}
\caption{Data sizes. N/A denotes no corresponding split; Builder candidates are gold / silver / rejected.}
\label{tab:supp-data}
\end{table}

\paragraph{Navigator data.}
Navigator queries are synthesized from source documents using an evidence-first procedure: supporting passages and atomic answer claims are selected before the question is generated. The teacher then interacts with the corresponding Wiki through the same search, read, link-following, stopping, and answering interface used at training time. For supervised data, we retain cleaned trajectories that recover all annotated evidence, discard tool failures and judge-rejected answers, remove repeated actions, and keep at most two trajectories per query. NAV-RL uses a separate query collection whose annotated evidence is reachable from the canonical training Wiki; policy trajectories are generated online rather than copied from teacher demonstrations.

\paragraph{Builder states and candidate generation.}
The teacher constructs a Wiki by processing ordered batches of source documents and emitting schema-constrained edit actions. Replaying the preceding construction events reconstructs the episode-specific pre-edit state $W_{\mathrm{before}}$ and binds it to the source batch consumed in that episode. Each candidate therefore contains the current sources, the corresponding Wiki snapshot, and the proposed structural patch. During BUILD-RL, the actor generates a new patch, which is applied to an isolated copy of $W_{\mathrm{before}}$ and discarded after scoring. This preserves a common pre-edit state for paired utility evaluation and prevents one candidate from altering later training instances.

\paragraph{Builder candidate filtering and quality tiers.}
Each synthetic BUILD candidate passes three deterministic screening layers before utility-based tiering. L0 checks schema validity and executability: paths must remain within approved top-level directories; \textsc{create} cannot overwrite an existing page; \textsc{update} must target an existing page; patches must be nonempty and untruncated; Wiki links must resolve; and a new page cannot exceed 0.75 line-level overlap with an existing page. L1 checks content quality and source support. An L1 failure rejects strict content-writing actions such as \textsc{write\_patchset}, but does not by itself reject non-content structural actions. L2 rejects structurally harmful changes, including increased orphaning or fragmentation.

When L3 evaluation is available, let
\[
\delta U_q = U_q(W_{\mathrm{after}})-U_q(W_{\mathrm{before}}).
\]
The frozen Navigator computes the mean change over each query set:
\begin{align}
\Delta U_{\mathrm{affected}}
&=\frac{1}{|Q_{\mathrm{affected}}|}
\sum_{q\in Q_{\mathrm{affected}}}\delta U_q,\\
\Delta U_{\mathrm{guard}}
&=\frac{1}{|Q_{\mathrm{guard}}|}
\sum_{q\in Q_{\mathrm{guard}}}\delta U_q.
\end{align}
A candidate that passes L0--L2 is \emph{gold} if $\Delta U_{\mathrm{affected}}\geq0.01$ and $\Delta U_{\mathrm{guard}}\geq-0.01$. A remaining candidate is \emph{silver} if both $\Delta U_{\mathrm{affected}}>-0.02$ and $\Delta U_{\mathrm{guard}}>-0.02$; all other candidates are rejected. An L0--L2 rejection overrides downstream utility. Valid \textsc{noop} and structural actions are assigned silver, as are valid content actions for which L3 is unavailable because no affected query exists or L3 evaluation is disabled.

\paragraph{Use across training stages.}
The quality tiers determine only which teacher trajectories enter BUILD-SFT. Rejected candidates are retained for filtering audits, not used as negative targets. Action-distribution controls retain all 628 gold and 2,343 of the 2,385 silver candidates, yielding 554 gold and 2,043 silver training instances and 74 gold and 300 silver validation instances. Joint SFT mixes these BUILD conversations with NAV supervision. NAV-RL uses its separate query collection and does not consume Builder tiers. BUILD-RL uses a separate collection of structurally executable episodes with valid pre-edit snapshots; tier labels are retained as metadata but do not determine rollout eligibility or reward. BUILD-RL and Joint RL generate patches online and score them with $R_{\mathrm{BUILD}}$.

Navigator records comprise 53.7\% Wiki-sufficient, 32.0\% digest-required, and 14.3\% article-required queries. Joint SFT samples NAV and BUILD examples at a $2{:}1$ ratio; because BUILD sequences are longer, the realized token distribution is approximately 36\% NAV and 64\% BUILD. Builder candidates are stratified by fan-in, author, and entity. Each instance receives at most three affected queries using page overlap, source overlap, semantic matching, and random fallback, in that order. Across 4,815 affected-query assignments, these stages contribute 758 (15.74\%), 3,408 (70.78\%), 455 (9.45\%), and 194 (4.03\%) assignments, respectively. Thus, 86.52\% of assignments have an exact page- or source-level correspondence, while only 4.03\% require deterministic same-author random completion. A stable hash shuffle selects two same-author guard queries with no source or page overlap. This fixed allocation balances query coverage against the cost of paired before/after evaluation and is shared by all Builder objectives.

We fix document and query partitions before generation. Training supervision, reward computation, filtering, calibration, and model selection use only the synthetic training and validation partitions. Test questions, answers, evidence annotations, and judgments are excluded from all of these stages. Exact-query overlap is zero across synthetic training, synthetic validation, and AuthTrace evaluation data, including between AuthTrace and NAV-SFT training data.

\textbf{Statistical testing.} Main-paper Table~3 and Supplementary Table~\ref{tab:supp-significance} report paired-bootstrap 95\% confidence intervals and paired-permutation $p$-values. Navigator and end-to-end outcomes are paired by AuthTrace query, whereas Builder outcomes are paired by evaluation instance. Holm correction is applied across the three comparisons in main-paper Table~3. In Supplementary Table~\ref{tab:supp-significance}, the two system comparisons are corrected within each AuthTrace stratum; the three Navigator metrics and four Builder comparisons form separate families. These tests condition on the trained policies and quantify uncertainty over the corresponding evaluation samples.

\section{Complete Experimental Results}

\subsection{Scale Transfer}

\begin{table}[!htbp]
\centering
\small
\setlength{\tabcolsep}{1.8pt}
\begin{tabular}{llcccc}
\toprule
Backbone & Method & Single & Low & High & All \\
\midrule
Qwen3.5-9B & Joint SFT & 65.8 & 49.3 & 39.8 & 58.2 \\
Qwen3.5-9B & \textbf{WikiLoop} & \textbf{69.1} & \textbf{54.8} & \textbf{47.5} & \textbf{62.6} \\
\midrule
Qwen3.5-35B-A3B & Joint SFT & 68.2 & 52.0 & 42.2 & 60.7 \\
Qwen3.5-35B-A3B & \textbf{WikiLoop} & \textbf{70.8} & \textbf{58.1} & \textbf{49.4} & \textbf{64.8} \\
\bottomrule
\end{tabular}
\caption{AuthTrace scale transfer with official stratum weighting.}
\label{tab:supp-scale}
\end{table}

\subsection{Judge Robustness}

To assess dependence on the primary DeepSeek-V4-Flash judge, we use GLM-5.1 to rescore all method outputs under the same AuthTrace rubric. Table~\ref{tab:supp-second-judge} reports AC with the official stratum weighting.

\begin{table}[!htbp]
\centering
\small
\setlength{\tabcolsep}{3.2pt}
\begin{tabular*}{\columnwidth}{@{\extracolsep{\fill}}lcccc@{}}
\toprule
Method & Single & Low & High & All \\
\midrule
None (Closed-book) & 4.2 & 6.3 & 5.1 & 4.9 \\
Vanilla RAG (BM25) & 38.7 & 16.5 & 11.4 & 29.5 \\
Vanilla RAG (Dense) & 44.6 & 15.8 & 14.9 & 33.3 \\
RAPTOR & 35.8 & 14.4 & 12.7 & 27.3 \\
LightRAG & 38.1 & 12.7 & 8.2 & 27.7 \\
HippoRAG~2 & 64.8 & 41.5 & 33.0 & 54.7 \\
LLM-Wiki, base & 63.8 & 49.5 & 43.0 & 57.4 \\
\textbf{WikiLoop} & \textbf{65.7} & \textbf{59.4} & \textbf{53.6} & \textbf{62.5} \\
\bottomrule
\end{tabular*}
\caption{AuthTrace AC under the GLM-5.1 judge.}
\label{tab:supp-second-judge}
\end{table}

WikiLoop ranks first in every stratum and overall under GLM-5.1. It exceeds LLM-Wiki, base by 1.9, 9.9, 10.6, and 5.1 points on Single, Low, High, and All, respectively, and exceeds HippoRAG~2 by 7.8 points on All. Although absolute scores and some baseline orderings vary between judges, the main pattern is stable: WikiLoop achieves its largest gains over LLM-Wiki, base on multi-document queries and retains the highest aggregate AC. Table~\ref{tab:supp-human-eval} further compares both automatic judges with human assessment.

\section{Additional Analyses}

\subsection{Efficiency Summary}

\begin{table}[!htbp]
\centering
\small
\begin{tabular}{lcc}
\toprule
Stage & Wall-Clock (h) & GPUs (Train / Eval.) \\
\midrule
Joint SFT & 4.2 & 16 / 0 \\
NAV-RL (Full) & 23.7 & 32 / 0 \\
BUILD-RL (Full) & 25.7 & 32 / 32 \\
Joint RL & 14.4 & 32 / 32 \\
\bottomrule
\end{tabular}
\caption{Training time and GPU allocation; evaluation GPUs serve the frozen Navigator.}
\label{tab:supp-training-cost}
\end{table}

Online latency is macro-averaged over the five author subsets. Offline compilation counts prompt and completion tokens used to construct a complete index or Wiki from the same raw corpora; WikiLoop uses the final shared policy's BUILD role. Training and online QA costs are excluded.

\begin{table}[!htbp]
\centering
\small
\begin{tabular*}{0.78\columnwidth}{@{\extracolsep{\fill}}lc@{}}
\toprule
Method & Online Latency (s/query) $\downarrow$ \\
\midrule
None (Closed-book)       & 10.35 \\
Vanilla RAG (BM25)       & \textbf{7.60} \\
Vanilla RAG (Dense)      & 16.36 \\
RAPTOR                   & 24.78 \\
LightRAG                 & 70.94 \\
HippoRAG~2               & 26.78 \\
LLM-Wiki, base           & 16.90 \\
\textbf{WikiLoop}        & \underline{15.76} \\
\bottomrule
\end{tabular*}
\vspace{3pt}

{\setlength{\tabcolsep}{1.8pt}
\begin{tabular*}{\columnwidth}{@{\extracolsep{\fill}}lcccccc@{}}
\multicolumn{7}{c}{Offline Compilation Tokens (M) $\downarrow$} \\
\cmidrule(lr){1-7}
Method & Lu & Zhou & Zhu & Xiao & Yu & Total \\
\midrule
RAPTOR & 0.44 & 0.40 & 0.30 & 0.10 & 0.20 & 1.44 \\
LightRAG & 25.03 & 85.24 & 58.21 & 31.31 & 43.76 & 243.55 \\
HippoRAG~2 & 10.77 & 5.90 & 4.48 & 2.72 & 4.62 & 28.49 \\
LLM-Wiki, base & 19.30 & 69.90 & 46.00 & 23.80 & 35.00 & 194.00 \\
\textbf{WikiLoop} & 18.64 & 66.27 & 43.88 & 23.11 & 33.25 & 185.15 \\
\bottomrule
\end{tabular*}
}
\caption{AuthTrace latency and offline construction cost. Boldface marks the overall latency minimum; underlining marks the minimum among structured retrieval systems.}
\label{tab:supp-efficiency}
\end{table}

WikiLoop averages 15.76 seconds per query, the lowest latency among the evaluated structured retrieval systems. Relative to LLM-Wiki, base, it reduces latency by 6.7\%, from 16.90 to 15.76 seconds, while increasing AuthTrace All AC from 56.3 to 62.6. It is also 36.4\% faster than RAPTOR, 41.2\% faster than HippoRAG~2, and 77.8\% faster than LightRAG.

BM25 is fastest overall at 7.60 seconds per query. Relative to BM25, WikiLoop improves All AC by 31.6 points at approximately 2.1 times the latency. Relative to HippoRAG~2, it improves All AC by 6.8 points while reducing latency by 41.2\%, providing a stronger accuracy--latency trade-off among the structured systems.

Reconstructing the complete Wiki with the final joint policy's BUILD role reduces compilation cost from 194.00M tokens for LLM-Wiki, base to 185.15M for WikiLoop, a reduction of 8.85M tokens, or 4.6\%. The reduction occurs across all five author corpora and ranges from 2.9\% to 5.2\%. Together with the increase in AuthTrace All AC from 56.3 to 62.6, this result indicates a better quality--construction-cost trade-off within the same Wiki-building framework.

WikiLoop also uses 24.0\% fewer compilation tokens than LightRAG (185.15M versus 243.55M), although RAPTOR and HippoRAG~2 use substantially fewer. These cross-paradigm totals provide system-level context; the most direct comparison is with LLM-Wiki, base, for which the learned Builder produces a more useful Wiki at a modestly lower full-rebuild token cost.

\subsection{Error Analysis}

\begin{figure}[!htbp]
\centering
\includegraphics[width=0.96\columnwidth]{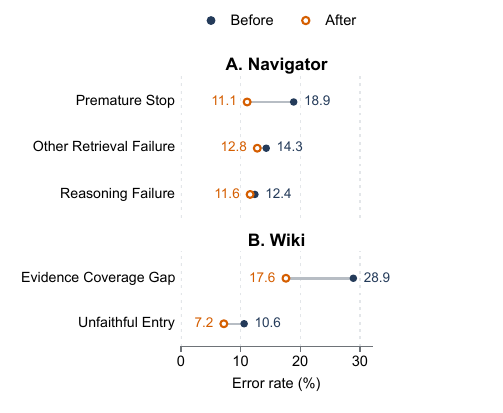}
\caption{Error rates before and after role-specific reinforcement learning. Panel A compares Joint SFT with NAV-RL (Full) over 2,099 queries: premature stopping ($ER<1$ with an active stop), other retrieval failure ($ER<1$ otherwise), and reasoning failure ($ER=1$, $AC=0$). Panel B reports paired errors before and after applying BUILD-RL (Full) patches to isolated Wiki copies. Filled and open markers indicate before and after; lower is better.}
\label{fig:supp-errors}
\end{figure}

Role-specific reinforcement learning changes the error profile in ways consistent with the two objectives. All Navigator rates use the same 2,099 evaluation queries as their denominator. The two incomplete-evidence categories sum to 33.2\% before NAV-RL and 23.9\% after NAV-RL, matching $100-ER\ \mathrm{Full}$ for the corresponding models in the main paper. Premature stopping shows the largest reduction, from 18.9\% to 11.1\%. Other retrieval failures decrease more modestly, from 14.3\% to 12.8\%, leaving page localization and evidence recovery as important residual bottlenecks. Reasoning failures decline only from 12.4\% to 11.6\%, indicating that NAV-RL primarily improves evidence acquisition rather than answer reasoning after full evidence recovery.

For the Builder, before and after statistics use the same affected query--patch pairs. Each valid patch is applied to an isolated copy of $W_{\mathrm{before}}$ to obtain $W_{\mathrm{after}}$ and discarded after scoring; both rates use the complete paired evaluation set as their denominator. Evidence coverage gaps decrease from 28.9\% to 17.6\%, while unfaithful entries decrease from 10.6\% to 7.2\%. The larger coverage improvement is consistent with optimizing edits through downstream navigation utility. Guarded utility-difference training also reduces unfaithful entries, although this error mode remains present.

\subsection{Human Evaluation of Judge Reliability}

We sample 200 AuthTrace queries: 100 Single, 60 Low, and 40 High. Evaluating outputs from WikiLoop, LLM-Wiki, base, and HippoRAG~2 yields 600 response items. Two annotators independently score every item using the same 0--3 AuthTrace rubric as the automatic judges. Method identities and automatic-judge scores are hidden, and the three outputs for each query are presented in randomized order. Disagreements are adjudicated after independent annotation to form the human consensus. Human All AC is reweighted to the official AuthTrace stratum proportions.

Table~\ref{tab:supp-human-eval} reports exact agreement and quadratically weighted Cohen's $\kappa$ before adjudication, together with human AC for the three principal systems. DeepSeek--human and GLM--human agreement are computed against the adjudicated consensus.

\begin{table}[!htbp]
\centering
\small
\begin{tabular*}{\columnwidth}{@{\extracolsep{\fill}}lcc@{}}
\multicolumn{3}{c}{\textbf{A. Judge Agreement}}\\[0.2ex]
\toprule
Comparison & Exact (\%) & $\kappa_w$ \\
\midrule
Human 1--Human 2 & \textbf{79.2} & \textbf{0.81} \\
DeepSeek--Human & 74.5 & 0.75 \\
GLM--Human & 72.8 & 0.72 \\
\bottomrule
\end{tabular*}

\vspace{4pt}

\begin{tabular}{lcccc}
\multicolumn{5}{c}{\textbf{B. Human AC}}\\[0.2ex]
\toprule
Method & Single & Low & High & All \\
\midrule
HippoRAG~2 & \textbf{67.7} & 39.4 & 30.8 & 55.6 \\
LLM-Wiki, base & 65.0 & 46.7 & 40.8 & 57.2 \\
\textbf{WikiLoop} & 67.3 & \textbf{57.8} & \textbf{50.8} & \textbf{62.8} \\
\bottomrule
\end{tabular}
\caption{Blinded human evaluation on 200 stratified AuthTrace queries. Panel A reports human--human agreement before adjudication and automatic-judge agreement with the adjudicated human consensus. Panel B reports consensus human AC, with All reweighted to the official stratum proportions.}
\label{tab:supp-human-eval}
\end{table}

\noindent\textbf{Results.}
Before adjudication, the two annotators obtain 79.2\% exact agreement and quadratically weighted Cohen's $\kappa=0.81$. Against the adjudicated human consensus, DeepSeek-V4-Flash achieves 74.5\% exact agreement and $\kappa_w=0.75$, while GLM-5.1 achieves 72.8\% exact agreement and $\kappa_w=0.72$. Most disagreements involve adjacent partial-credit categories rather than reversals between fully correct and incorrect judgments.

Human evaluation preserves the main system-level pattern. HippoRAG~2 retains a marginal advantage on Single-document queries (67.7 versus 67.3), whereas WikiLoop achieves the highest AC on Low and High multi-document queries. After reweighting to the official AuthTrace distribution, WikiLoop reaches 62.8 All AC, exceeding LLM-Wiki, base by 5.6 points and HippoRAG~2 by 7.2 points. The advantage on multi-document evidence synthesis therefore remains under blinded human assessment and is not specific to either automatic judge.

\subsection{Significance of Primary Comparisons}

Table~\ref{tab:supp-significance} reports paired tests for the primary system and role-objective comparisons. End-to-end tests use all 2,099 AuthTrace queries for All and the corresponding 1,285 Single, 552 Low, and 262 High queries for each stratum. Builder tests use matched evaluation instances.

\begin{table}[!htbp]
\centering
\small
\setlength{\tabcolsep}{1.8pt}
\begin{tabular*}{\columnwidth}{@{\extracolsep{\fill}}llcc@{}}
\multicolumn{4}{c}{\textbf{A. End-to-End Systems}}\\[0.2ex]
\toprule
Reference & Stratum & $\Delta$ (95\% CI) & Holm $p$ \\
\midrule
LLM-Wiki, base & All    & $+6.3\;[+5.1,+7.5]$   & $<0.001$ \\
               & Single & $+2.4\;[+0.8,+4.0]$   & $0.006$ \\
               & Low    & $+12.8\;[+10.2,+15.5]$ & $<0.001$ \\
               & High   & $+12.3\;[+8.2,+16.4]$ & $<0.001$ \\
HippoRAG~2     & All    & $+6.8\;[+5.5,+8.1]$   & $<0.001$ \\
               & Single & $-0.7\;[-2.5,+1.1]$   & $0.440$ \\
               & Low    & $+18.3\;[+15.3,+21.3]$ & $<0.001$ \\
               & High   & $+19.5\;[+15.0,+24.0]$ & $<0.001$ \\
\bottomrule
\end{tabular*}

\vspace{4pt}

\begin{tabular*}{\columnwidth}{@{\extracolsep{\fill}}llcc@{}}
\multicolumn{4}{c}{\textbf{B. Navigator: Full $-$ Ungated}}\\[0.2ex]
\toprule
Comparison & Metric & $\Delta$ (95\% CI) & Holm $p$ \\
\midrule
Full $-$ Ungated & AC & $+2.3\;[+1.2,+3.4]$ & $<0.001$ \\
                 & ER Full & $+12.1\;[+10.0,+14.2]$ & $<0.001$ \\
                 & Prem. Stop & $-13.1\;[-15.3,-10.9]$ & $<0.001$ \\
\bottomrule
\end{tabular*}

\vspace{4pt}

\begin{tabular*}{\columnwidth}{@{\extracolsep{\fill}}llcc@{}}
\multicolumn{4}{c}{\textbf{C. Builder Objectives}}\\[0.2ex]
\toprule
Contrast & Metric & $\Delta$ [95\% CI] & $p_H$ \\
\midrule
NoG. $-$ Abs. & Held-out & $+0.043\,[+0.030,+0.056]$ & $<0.001$ \\
               & Guard Reg. & $+0.021\,[+0.001,+0.041]$ & $0.078$ \\
Full $-$ NoG. & Held-out & $+0.007\,[-0.002,+0.016]$ & $0.120$ \\
               & Guard Reg. & $-0.030\,[-0.047,-0.013]$ & $0.0036$ \\
\bottomrule
\end{tabular*}
\caption{Paired tests for main-paper Tables~1 and~2. In Panel A, $\Delta$ is WikiLoop minus the reference; in Panels B and C, it follows the stated contrast. Abs., NoG., Held-out, and Guard Reg. denote absolute post-edit utility, utility difference without guard protection, held-out-Navigator $\Delta U$, and Guard Regression. Intervals are paired-bootstrap 95\% CIs; $p_H$ denotes a Holm-adjusted paired-permutation $p$-value.}
\label{tab:supp-significance}
\end{table}

System-level gains are statistically reliable except for the Single-stratum comparison with HippoRAG~2. Sufficiency gating improves all three primary Navigator outcomes. For the Builder, marginal utility improves held-out utility, and guard protection reduces regression. The smaller increase in Guard Regression from absolute to unguarded marginal utility and the held-out utility change from the unguarded to the full objective are not reliable after correction.

\end{document}